\documentclass[10pt,twocolumn,letterpaper]{article}

\usepackage[pagenumbers]{iccv} 
\usepackage{multirow}
\usepackage[ruled]{algorithm2e}

\definecolor{iccvblue}{rgb}{0.21,0.49,0.74}
\usepackage[pagebackref,breaklinks,colorlinks,allcolors=iccvblue]{hyperref}

\def\paperID{14183} 
\def\confName{ICCV}
\def\confYear{2025}

\title{DiTFastAttnV2: Head-wise Attention Compression for \\ Multi-Modality Diffusion Transformers}

\author{Hanling Zhang\thanks{Equal contribution. $^\dag$ Corresponding author.}~$^{1,2}$ \quad Rundong Su$^{*3}$ \quad
Zhihang Yuan$^{\dag 1,2}$ \quad
Pengtao Chen$^{3}$ \quad
Mingzhu Shen$^{4}$ \quad \\
Yibo Fan$^{3}$ \quad
Shengen Yan$^{2}$ \quad
Guohao Dai$^{2,5}$ \quad
Yu Wang$^{1}$\\
\\
\fontsize{9pt}{15pt}\selectfont$^1$Tsinghua University\quad$^2$Infinigence AI\quad$^3$Fudan University\quad
$^4$Imperial College London\quad$^5$Shanghai Jiao Tong University
}

\begin{document}

\maketitle

\begin{abstract}
Text-to-image generation models, especially Multimodal Diffusion Transformers (MMDiT), have shown remarkable progress in generating high-quality images. However, these models often face significant computational bottlenecks, particularly in attention mechanisms, which hinder their scalability and efficiency. In this paper, we introduce DiTFastAttnV2, a post-training compression method designed to accelerate attention in MMDiT. Through an in-depth analysis of MMDiT’s attention patterns, we identify key differences from prior DiT-based methods and propose head-wise arrow attention and caching mechanisms to dynamically adjust attention heads, effectively bridging this gap. We also design an Efficient Fused Kernel for further acceleration. By leveraging local metric methods and optimization techniques, our approach significantly reduces the search time for optimal compression schemes to just minutes while maintaining generation quality. Furthermore, with the customized kernel, DiTFastAttnV2 achieves a 68\% reduction in attention FLOPs and 1.5$\times$ end-to-end speedup on 2K image generation without compromising visual fidelity.
\end{abstract}    
\section{Introduction}
\label{sec:intro}

\begin{figure}[]
    \includegraphics[width=1\linewidth]{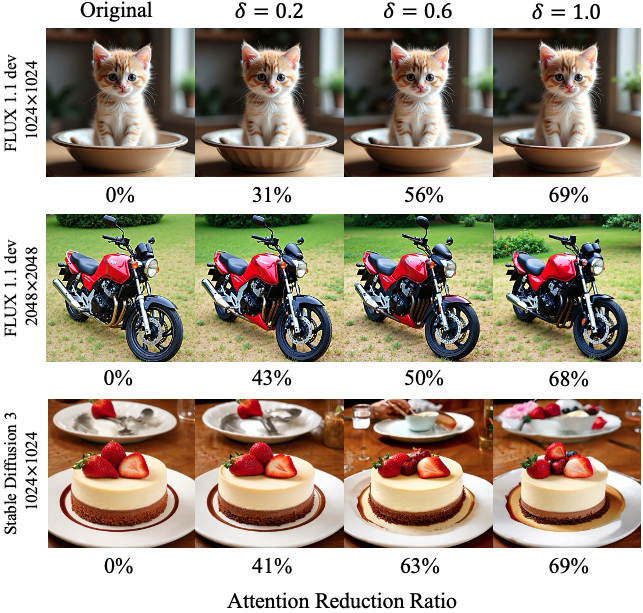}
    \caption{Visual comparison of image generation results using DiTFastAttnV2 under different compression threshold $\delta$. The percentage under each image stands for the attention reduction ratios under the setting. Our method achieves up to 69\% reduction in attention computation while preserving high-quality generation outputs. The visual fidelity remains consistent even at higher compression rates.}
    \label{fig:difav2_results}
\end{figure}



Rapid advancements in diffusion transformer models~\cite{chen2024pixart, esser2024scaling, videoworldsimulators2024,kong2024hunyuanvideo} have demonstrated impressive generative capabilities. As these models continue to scale up, inference efficiency emerges as a critical challenge, particularly in high-resolution image and long video generation scenarios. The computational complexity of attention mechanisms represents a significant bottleneck, necessitating attention compression techniques. While DiTFastAttn~\cite{yuan2024ditfastattn} has attempted to address this by identifying multi-dimensional redundancies in DiT models through sliding window attention and output caching, the adoption of such optimizations in mainstream architectures MMDiT~\cite{esser2024scaling} encounters fundamental limitations. These challenges stem from three underexplored dimensions that fundamentally constrain current acceleration approaches.

\begin{itemize}
    \item \textbf{Cross-Modal Attention Pattern Complexity}: Unlike previous diffusion transformers that only have image tokens in the self-attention module and use cross-attention for text modalities, MMDiTs have joint self-attention for both text and image tokens, creating a complex attention landscape. Our analysis reveals that visual tokens exhibit distinct attention characteristics, notably a strong diagonal locality pattern that remains consistent across different prompts. Conversely, language token interactions demonstrate significant semantic variability. Existing methods like DiTFastAttn, designed for uniform sliding window attention, fail to capture this nuanced attention dynamics. Simply applying standard sparse attention mechanisms risks truncating critical text-related information and disrupting the delicate multimodal alignment.

    \item \textbf{Head-wise Redundancy Heterogeneity}: Our detailed investigation uncovers heterogeneity in attention head behaviors across timesteps and spatial dimensions. We observed that attention heads within the same layer exhibit dramatically different characteristics: some heads demonstrate near-global attention, while others display highly localized patterns. Moreover, the redundancy across diffusion timesteps varies significantly between heads. Thus, existing layer-wise caching or sparse attention strategies that treat all heads uniformly risk eliminating crucial information encoded in rapidly evolving heads, thereby compromising generation quality.

    \item \textbf{Prohibitive Compression Plan Search Cost}: Existing compression schemes search~\citep{yuan2024ditfastattn} rely on exhaustive search algorithms that require more than 10 hours for compression plan generation, primarily due to compression evaluations through full denoising processes.
\end{itemize}

To address these challenges, we propose a comprehensive framework that tackles mismatches and heterogeneity in MMDiTs attention computation and enhances search strategies for model compression configuration. Our approach leverages head-wise attention patterns coupled with head-wise caching mechanism to form a fine-grained compression plan, preserving critical information while reducing computational redundancy. To reduce the cost of model configuration search, we develop a novel calibration approach using a single-layer Relative Squared Error (RSE) metric and headwise compression plan optimization, significantly reducing search time while maintaining generation performance. 

In summary, our contributions are threefold.

\begin{itemize}
    \item We propose a novel approach that integrates \textbf{head-wise attention} pattern adjustments, dynamic head-wise caching accompanied with an \textbf{efficient fused kernel} to optimize attention mechanisms in Multi-Modality Diffusion Transformers (MMDiT). This combination effectively resolves mismatches in window attention patterns and enhances computational efficiency by dynamically adjusting attention heads and fusing kernels for faster execution.
    
    \item We introduce a combination of \textbf{local calibration} strategy combined with \textbf{headwise compression plan optimization} to reduce the search cost in assigning configuration significantly without compromising visual fidelity. This targeted exploration of network structures enables faster and more efficient optimization, ensuring high performance while reducing search time.

    \item On 2K image generation task our method can reduce up to \textbf{68}\% attention computation and achieve \textbf{1.5$\times$} end-to-end speedup without image quality degradation. 
\end{itemize}

\section{Related Work}

\subsection{Diffusion Model}
\begin{figure}[t]
    \centering
    \includegraphics[width=0.9\linewidth]{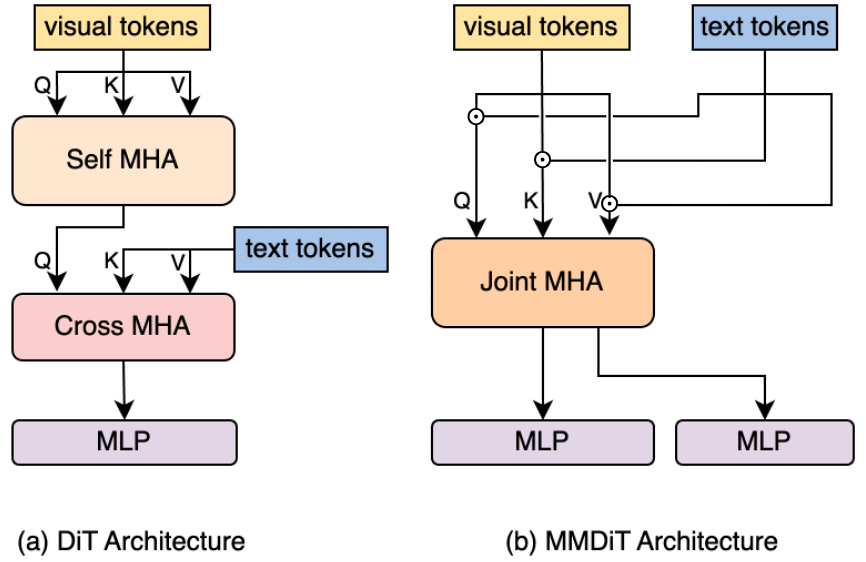}
    \caption{\textbf{DiT and MMDiT block architecture}. In MMDiT, after projections, visual and text tokens are concatenated for a joint self-attention.}
    \label{fig:model_arch}
\end{figure}

In recent years, diffusion models~\cite{rombach2022high} have achieved groundbreaking progress and are playing a vital role across various fields~\cite{dhariwal2021diffusion}, including image~\cite{song2020score, song2019generative}, video~\cite{blattmann2023stable}, and 3D generation~\cite{voleti2025sv3d}. Earlier diffusion models were primarily based on the UNet architecture; however, it has gradually been recognized that the Diffusion Transformer architecture~\cite{peebles2023scalable}, which follows scaling laws, offers greater advantages. This can be seen in notable models such as the Pixart series~\cite{chen2023pixart, chen2024delta} and Hunyuan-DiT in image generation, as well as in video generation with models like Sora~\cite{videoworldsimulators2024}, Open-Sora~\cite{zheng2024open}, and LTX-Video~\cite{hacohen2024ltx}. Proposed by SD3~\cite{esser2024scaling}, the MMDiT (Multimodal Diffusion Transformer) architecture has gradually replaced vanilla DiT architecture as the mainstream approach primarily due to its superior instruction following capabilities. Unlike previous Diffusion Transformers (DiT), as shown in Fig~\ref{fig:model_arch}, MMDiT eliminates the cross-attention module in DiT; instead, visual inputs and text embeddings are independently projected and then concatenated together for self-attention processing. Notable models adopting the MMDiT architecture include SD3~\cite{esser2024scaling} and FLUX~\cite{blackforestlabs_flux_2024} in image generation, as well as CogVideoX~\cite{yang2024cogvideox}, HunyuanVideo~\cite{kong2024hunyuanvideo}, and Mochi-1~\cite{genmo2024mochi} in video generation domains. This work is specifically focused on attention compression for MMDiTs.

\subsection{Efficient Diffusion Model}
Given the challenges of real-time generation in diffusion models, many researchers have explored acceleration techniques. Some approaches focus on pruning the model to decrease the parameter in noise estimation networks~\cite{zhang2024laptop, fang2023structuralpruningdiffusionmodels, castells2024ld}, while others use quantization techniques~\cite{shang2023post, li2023q, he2024ptqd} to reduce model size. Distillation methods have also been employed, either to compress noise estimation networks or to minimize the number of denoising steps required~\cite{kim2023architectural, salimans2022progressive}. Further, some studies aim to reduce the computational load by decreasing the token count during inference~\cite{bolya2023token, kim2024token, smith2024todo}, and others reuse features in the network between timesteps to speed up inference~\cite{ma2024deepcache, so2023frdiff, chen2024delta, wimbauer2024cache, li2023faster, zhang2024cross, selvaraju2024fora}. Our work is different: we consider the efficiency of diffusion models from an attention perspective.


\subsection{Efficent Attention Method}
In both Vision Transformers (ViTs~\cite{dosovitskiy2020vit}) and Large Language Models (LLMs~\cite{achiam2023gpt}), prior research has shown that attention computation displays inherent sparsity~\cite{wen2016sparsity}. Leveraging this property, various approaches have been proposed to accelerate Transformer inference through sparse computation. For ViTs, locality-aware mechanisms such as neighbor attention~\cite{beltagy2020longformer,hassani2023neighborhood} and stride attention~\cite{child2019stride} have been introduced. Additionally, the Swin Transformer~\cite{liu2021swin} employs a shifted window attention mechanism to limit computations within localized regions, effectively reducing attention latency. In the context of LLMs, techniques like StreamingLLM~\cite{xiao2023efficient} and DuoAttention~\cite{xiao2024duoattention} have identified the existence of an Attention Sink, leading to the development of novel attention patterns aimed at mitigating its impact. This has resulted in the introduction of windowed attention variants that enhance inference efficiency. For long-context LLMs, research~\cite{jiang2024minference} has identified three main attention patterns: A-shape, Vertical-Slash, and Block-Sparse. An adaptive sparse allocation strategy has also been proposed. This strategy dynamically assigns the optimal pattern configuration to each attention, facilitating more efficient sparse inference. For diffusion models, CLEAR~\cite{liu2024clear} has investigated the locality characteristics in DiT-based generation and introduced a circular attention mechanism that enables more efficient generation through distillation. In contrast to these approaches, our method seeks to optimize DiT inference efficiency through sparsity. Specifically, we introduce fine-grained head-level sparsity, which allows for more efficient computation while achieving superior optimization performance compared to existing methods.

\subsection{DiTFastAttn Framework}
In the DiTFastAttn~\cite{yuan2024ditfastattn}, the attention maps in DiT exhibit a diagonal pattern, with similar attention outputs between adjacent timesteps and similarities between the outputs of the CFG~\cite{ho2022classifier}. Based on these three insights, DiTFastAttn proposes three corresponding optimizations: Windowed Attention with Residual Sharing for attention, Attention Sharing across Timesteps, and Attention Sharing across CFGs. These methods are implemented through a greedy search algorithm to assign the most suitable strategy to each attention mechanism. While these approaches yield notable improvements in efficiency and generation quality, some aspects have yet to be explored: 
(1) DiTFastAttn employs a unified acceleration strategy across all attention heads. However, the sparsity patterns of different heads exhibit inherent variations. As a result, applying a one-size-fits-all approach may lead to suboptimal efficiency and performance.
(2) As diffusion models rapidly evolve, MMDiT models~\cite{esser2024scaling, blackforestlabs_flux_2024} have been demonstrated to yield superior generation quality. MMDiT introduces a novel attention mechanism that simultaneously handles text and image tokens, eliminating the need for CFG operations~\cite{blackforestlabs_flux_2024} and introduce multi-modality in self attention computation, which makes the windowed attention and CFG-based attention sharing techniques in DiTFastAttn incompatible; and (3) Compression plan search algorithm in DiTFastAttn suffers from high computational complexity and incurs considerable overhead. To comprehensively address these issues, we introduce DiTFastAttnV2, which is designed to accommodate more advanced architectures, achieve greater acceleration, and enhance optimization efficiency.

\section{Method}
\label{sec:method}

\begin{figure}[h]
    \centering
    \includegraphics[width=0.9\linewidth]{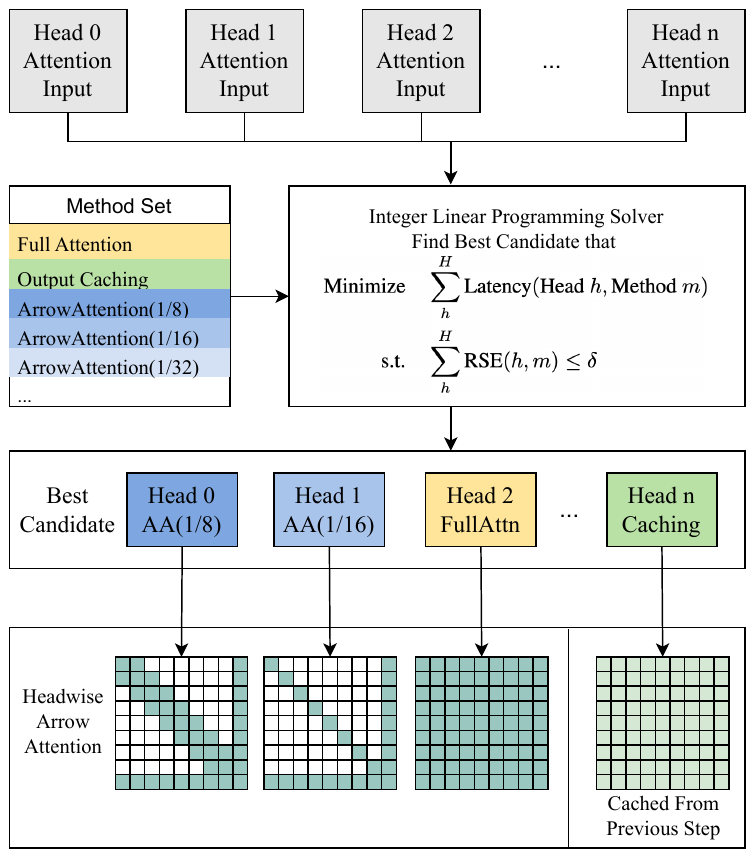}
    \caption{Overview of DiTFastAttnV2}
    \label{fig:difav2_method}
\end{figure}


\subsection{Head-wise Arrow Attention}

Unlike DiT, which employs cross-attention mechanisms to incorporate text guidance, our proposed MMDiT implements a joint attention architecture. In this structure, visual and language tokens are concatenated before applying self-attention operations. The length of visual tokens is variable and scales with the resolution of the generated image, while language tokens maintain a fixed length. The attention map of the joint attention can be partitioned into four areas representing: visual-to-visual token interactions, visual-to-language token interactions, language-to-visual token interactions, and the interaction between language tokens. To systematically evaluate the impact of this architectural design on attention, we first conduct a comprehensive analysis of MMDiT attention patterns. Our analysis reveals the following findings:


It is observed that a considerable part of the attention map shows an obvious locality in the visual-to-visual interaction area, and the attention score is mainly concentrated in a certain range around the diagonal line. This locality pattern remains largely invariant across different prompts, suggesting it represents an intrinsic property of the architecture rather than a prompt-dependent feature.


The locality of the visual part exhibits significant heterogeneity. The distribution of attention patterns varies considerably, with distinct patterns emerging not only between different layers but also among attention heads within the same layer. As illustrated in the figure, in the given layer, some heads exhibit nearly full attention and others show local attention patterns. Furthermore, among the heads exhibiting localized attention, the size of the high-attention concentration regions varies substantially.


The remaining part involving language token interactions demonstrates substantial variability across different text prompts. These regions exhibit no consistent or predictable patterns, suggesting that these attention mechanisms dynamically adapt to the specific semantic content of each prompt. 


\begin{figure*}[]
    \centering
    \includegraphics[width=0.9\linewidth]{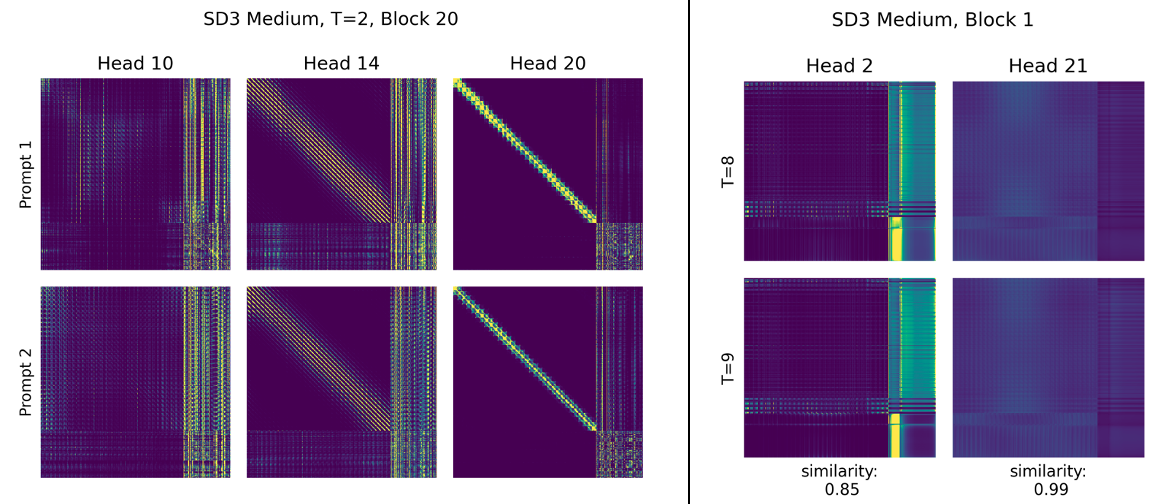}
    \caption{\textbf{Left}: Selected head attention map examples. Head 10 exhibits global attention in the visual-visual token interaction area, while heads 14 and 20 exhibit different extents of local attention patterns. In text interaction areas, attention varies with different prompts. \textbf{Right}: Different heads within a layer exhibit different levels of redundancy across denoising steps. The attention similarity between adjacent time steps in Head 21 is significantly higher than that in Head 2.}
    \label{fig:arrow_analysis}
\end{figure*}

Based on the above findings, we propose an attention tailored to the sparse nature of MMDiT. In the visual part, local attention is used to discard non-contributing tokens outside the neighborhood. Conversely, for regions involving text token interactions (visual-to-language, language-to-visual, and language-to-language), we preserve all these attention without compression, maintaining their full representational capacity. Such a pattern visually resembles an arrow, so we named this pattern arrow attention. To accommodate the heterogeneous distribution of locality characteristics observed across different heads and layers, we further introduce a mixed attention design where each attention head can select between full attention and Arrow Attention.

\subsection{Head-wise Caching}


Caching methods leverage timestep redundancy in diffusion models to reduce attention computational costs. However, does this type of redundancy exhibit heterogeneity among different attention heads, similar to the spatial redundancy we observed in MMDiT? To address this question, we conducted a fine-grained analysis of attention similarity between consecutive diffusion steps.
Our findings reveal significant head-specific variations between adjacent timesteps within the same transformer block. Some heads exhibit substantial changes while others remain relatively stable across timesteps. This heterogeneity suggests that naive layer-wise caching approaches, which treat all heads identically, risk eliminating critical information encoded in rapidly evolving heads, potentially compromising the quality of generated outputs.


To address this problem, we propose to perform head-wise caching on the attention output and skip the attention calculation for heads with high similarity in adjacent time steps.

\subsection{Fused Kernel for Multi-Strategy Attention}


We designed a specialized fused kernel that efficiently integrates Head-wise Arrow Attention with Head-wise Caching mechanisms. This custom operator allows each attention head to independently select from three operational modes: full attention computation, calculation skipping (with cached result reuse), or arrow attention with user-specified window size. After primary computation, the heads that opted to skip calculations are populated with their corresponding cached attention outputs.

\begin{figure}[h]
    \centering
    \includegraphics[width=0.9\linewidth]{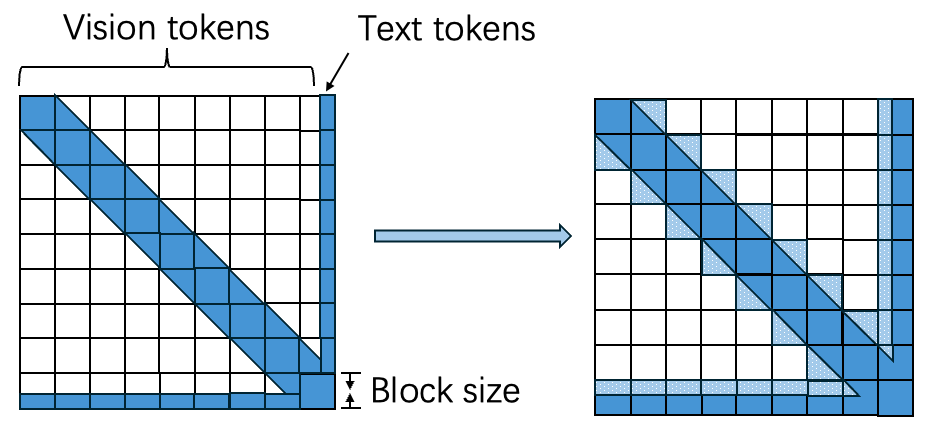}
    \caption{Visualization of arrow attention maps in our efficient kernel implementation. The kernel operates in a block sparse way, transforming mix blocks into dense blocks to reduce memory access overhead.}
    \label{fig:arr_attn_map}
\end{figure}

Our implementation incorporates these heterogeneous attention patterns while maintaining computational efficiency. As shown in Fig.~\ref{fig:arr_attn_map}, for the local attention regions in arrow attention, we implemented block-based sparsity patterns that ensure each computational block is dense, thus minimizing overhead from irregular memory access patterns and maximizing throughput on modern GPU architectures.

\subsection{Efficient Compression Plan Search}

\subsubsection{Efficient Calibration Metric}


In previous work, the MSE of the final output was used as a metric. Such a metric can intuitively reflect the impact of the applied acceleration method on the final output of the model, but the profiling speed is very slow. Taking the same comprehensive acceleration solution DiTFastAttn as an example, it takes more than 10 hours to generate an acceleration solution for 50 timesteps 2K image generation using FLUX. Extending to headwise, using the final output to generate an acceleration solution requires ($T \times L \times M \times H$) complete inferences. For larger models, it means a calibration cost more than 200 hours on a Nvidia A100 card. Therefore, we need a more efficient metric for compression method search. We hope to explore using only a single-layer output metric to calibrate the model. For the headwise method search we designed, we found that using a single-layer rse can ideally reflect the actual impact of each method. While ensuring the quality of generation, we can control the calibration cost to ($T\times M$), that is, $M$ normal inferences, allowing users to generate compression solutions more flexibly for their own settings. In this work, for each head, we use the relative squared error between attention output after applying certain method $y_{m}$ and the original output $y_{o}$ as the metric that reflects the influence $\mathcal{I}$ of applying the method.

\begin{equation}
\begin{aligned}
\mathcal{I}(m) = \frac{\sum (y_{m} - \bar{y}_o)^2}{\sum (y_o - \bar{y}_o)^2}
\end{aligned}
\end{equation}

\subsubsection{Head-wise Compression Plan Optimization}


Analyzing the dependencies in the compression plan search process reveals complex interdependencies across both temporal and network dimensions. The output of each model layer demonstrates dependencies that span across timesteps and attention layers. Importantly, compression decisions made for earlier timesteps fundamentally alter the input conditions for subsequent timesteps. Similarly, compression plans established for shallow attention layers directly influence the optimal compression strategies for deeper layers. Based on these observed interdependencies, we implement a progressive updating approach that systematically searches and refines the compression plan in a timestep-by-timestep and layer-by-layer manner. 

For each specific layer at a specific timestep, our objective is to identify a compression configuration that maximizes computational efficiency while maintaining generation quality above a predefined threshold. Since attention heads within each layer operate independently, we formulate the compression plan determination as an integer optimization problem. This approach enables us to derive the optimal acceleration scheme while minimizing a given Relative Squared Error (RSE) budget.

However, our empirical investigations revealed an important limitation: under fixed influence budgets, layers with lower attention redundancy tend to concentrate the entire influence budget on a single head, potentially causing critical information loss. To address this issue, we introduce a constraint coefficient
c that imposes head-specific limitations on compression influence. This coefficient prevents any single head from absorbing excessive compression budget, ensuring a more balanced distribution across heads and preserving the model's representational diversity.

For head $h$, and number of head $n$, given $m \in$ method, the candidate set $M$. Denote whether a method $m$ is selected for head $h$ as a binary variable $X_{h,m} \in \{0, 1\}$. Denote the latency of applying a method $m$ on an attention head as $\mathcal{L}_{m}$. Denote the influence of using a method $m$ on the head $h$ as $\mathcal{I}(h, m)$. For user-set threshold $\delta$, head constraint coefficient $c$ the optimization problem can be formed as following: 
\begin{equation}
\begin{aligned}
\min \quad & \sum_{h}^{H}\sum_{m}^{M}\mathcal{L}(h,m)\\
\textrm{s.t.} \quad & \sum_{h}^{H}\sum_{m}^{M}X_{h,m}\mathcal{I}(h,m) \leq \delta \quad 
 \forall h \in H\\
\quad & \mathcal{I}(h,m) \leq \frac{c}{n}\delta \\
  & \sum_{m}^{M} X_{h,m} \leq 1 \quad 
 \forall h \in H\\
\end{aligned}
\end{equation}


\begin{algorithm}
    \SetKwInOut{Input}{Input}

    \Input{query $Q$, key $K$, value $V$, Method Set $\mathcal{M}$, Threshold $\delta$, Number of Head H}
    
    $O$ $\leftarrow \text{Attention}(Q, K,V)$ \\
        \For{\textnormal{m} $\in$ $\mathcal{M}$}{
        $O'$ $\leftarrow \text{Attention}_{m}(Q, K,V)$ \\
            \For{\textnormal{head} $h$ 
            \textnormal{in} $H$}{
                \textnormal{Influence}$[h,m]\leftarrow RSE(O[h], O'[h]) $
            }
        }

    \textnormal{Optimized Plan} $\leftarrow$ \textnormal{ILPSolver(Influence, $\delta$)} \\
    $O$ $\leftarrow \text{HeadwiseAttention}(Q,K,V,\textnormal{Optimized Plan})$ \\
    idx $\leftarrow$ head Index of output caching head in \textnormal{Optimized Plan} \\
    $O[idx] = O_{cache}[idx]$

    \Return{} $O$
    \caption{Single Block Calibration Strategy}
    \label{alg:greedy}
\end{algorithm}



%


\section{Experiments}

\begin{table*}[t]
\centering
\caption{Quantitative Result of DiTFastAttnV2 on Stable Diffusion 3 and FLUX.1 - dev}
\label{tab:quantitative_result}
\begin{tabular}{lccccccc}
\toprule
Model &
  Resolution &
  Settings &
  \begin{tabular}[c]{@{}l@{}}Attention \\ Sparsity\end{tabular} &
  LPIPS &
  SSIM &
  HPSv2 &
  CLIP \\ \hline
\multirow{4}{*}{Stable Diffusion 3} &
  \multirow{4}{*}{1024} &
  Original &
  0 &
  - &
  - &
  0.2926 &
  0.3254 \\
 &                       & $\delta = 0.2$ & 0.41 & 0.182 & 0.716 & 0.2933 & 0.3251 \\
 &                       & $\delta = 0.6$ & 0.63 & 0.266 & 0.616 & 0.2933 & 0.3246 \\
 &                       & $\delta = 1.0$ & 0.69 & 0.318 & 0.560 & 0.2923 & 0.3245 \\ \hline
\multirow{8}{*}{FLUX.1 - dev} &
  \multirow{4}{*}{1024} &
  Original &
  0 & - & - & 0.2949 & 0.3172 \\
 &                       & $\delta = 0.2$ & 0.31 & 0.142 & 0.760 & 0.2959 & 0.3170 \\
 &                       & $\delta = 0.6$ & 0.56 & 0.227 & 0.644 & 0.2959 & 0.3176 \\
 &                       & $\delta = 1.0$ & 0.69 & 0.267 & 0.565 & 0.2952 & 0.3178 \\ \cline{2-8} 
 & \multirow{4}{*}{2048} & Original       & 0    & -     & -     &  0.2862  &   0.3169   \\
 &                       & $\delta = 0.2$ & 0.43 & 0.242    &    0.646   &  0.2883  &   0.3164    \\
 &                       & $\delta = 0.6$ & 0.50 &  0.343     &   0.539    &  0.2886    &    0.3159    \\
 &                       & $\delta = 1.0$ & 0.68 &  0.393   &   0.497   &   0.2852   &    0.3163    \\ \hline
\end{tabular}
\end{table*}

\subsection{Settings}

 We evaluate the performance of DiTFastAttnV2 on two mainstream opensourced generative models: Stable Diffusion 3 and FLUX dev 1.1. For each model, we generated 5,000 images using MS-COCO 2014 captions at resolutions of $1024 \times 1024$ and $2048 \times  2048$ pixels. We set the number of inference steps to 50 to ensure the generative quality of the original model and default values for all other parameters. We selected a comprehensive set of metrics for evaluation, including Structural Similarity Index Measure (SSIM), Learned Perceptual Image Patch Similarity (LPIPS)~\cite{zhang2018unreasonable}, Human Perception Score version 2 (HPSv2)~\cite{wu2023human}, CLIP score~\cite{radford2021learning}, and Inception Score~\cite{salimans2016improved}.

\subsection{Generation Results}


We present the evaluation results of DiTFastAttnV2 under different thresholds along with the corresponding reduction in attention FLOPs. At configurations of $\delta=0.2$ and $\delta=0.6$, the generated results maintain a comparable high similarity with the original images while achieving higher HPSv2 scores and comparable CLIP scores, indicating that our compression method does not degrade the quality of image generation. Even at higher compression rates, although the similarity between the generated images and the original ones decreases, our method is still able to maintain HPSv2 and CLIP scores that are comparable to those of the original generated images.

Fig.~\ref{fig:difav2_results} presents image generation samples in different threshold configurations. Under $\delta=0.2$ configuration, the generation images maintain a high degree of similarity to the original outputs while achieving up to 41\% attention reduction. Under higher compression settings ($\delta = 0.4$, $\delta=0.6$), the generated images exhibit differences in details and backgrounds but still ensure that the main subjects and perspectives of the generated images remain consistent with the original.

\subsection{Comparison with Existing Attention Compression Methods}

\begin{figure}[]
    \includegraphics[width=1\linewidth]{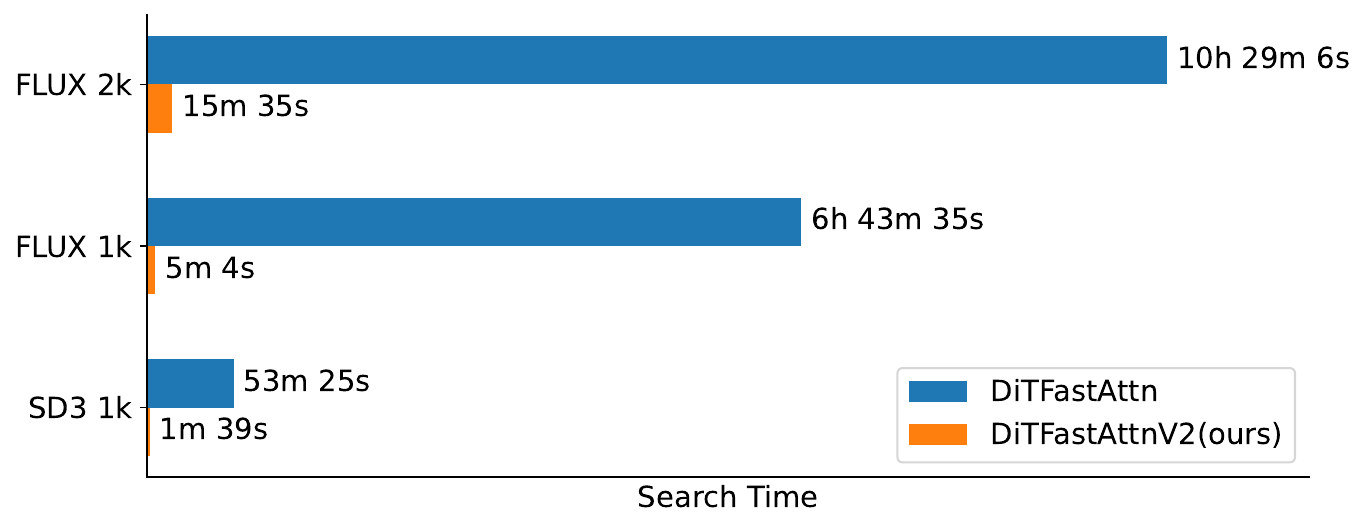}
    \caption{Compression Plan Search Time.}
    \label{fig:search_time_comparison}
\end{figure}


We compared our method with efficient attention algorithm DiTFastAttn. Considering that sliding window attention that DiTFastAttn adopts might not correctly capture the text information in MMDiT, we implement a variant of DiTFastAttn that uses ArrowAttention we proposed to replace sliding window attention that keeps text information. Caching by layer is shown as a simple but effective method in previous work, so here we implement an attention caching baseline that reuses previous attention output for every two steps.


With a large threshold setting, DiTFastAttn and its variants still fail to achieve a high level of attention reduction, which validates the previously mentioned low consistency of the same attention layer pattern in MMDiT. Simple attention caching, while able to maintain the semantic information of the generated images at a 50\% reduction, still results in a decrease in human preference scores and a significant difference from the original images. In contrast, our model maintains semantic information and aesthetic quality comparable to the original images and is more similar to the original images while reducing attention by 55\%. Even when the attention rate is increased to 63\%, DiTFastAttnV2 still retains considerable generation results.


While reviewing the generated images, we found that the basic caching method, due to its heavy dependence on earlier time steps, occasionally leads to incomplete object generation. In contrast, DiTFastAttn displays artifacts in some scenarios. Our method takes full advantage of the redundancy available across both time steps and spatial dimensions, preserving the image composition even at elevated compression rates, without diminishing the integrity and visual quality of the output.

\begin{table*}[h]
\caption{Quantitative Result of DiTFastAttnV2 Comparing with Other Attention Compression Method on Stable Diffusion 3 }
\label{tab:result_comparison_sd3}
\centering
\begin{tabular}{lcccccc}
\toprule
Method                                           & \begin{tabular}[c]{@{}l@{}}Attention \\ Sparsity\end{tabular} & LPIPS          & SSIM           & HPSv2           & CLIP            & IS              \\
\midrule
Original Stable Diffusion 3                & 0                  & -              & -              & 0.2926          & 0.3254          & 22.346          \\
DiTFastAttn                                      & 0.32               & 0.301          & 0.607          & 0.2839          & 0.3229          & 21.185          \\
DiTFastAttn with Arrow Attention                 & 0.44               & 0.297          & 0.616          & 0.2841          & 0.3219          & 21.494          \\
Attention Caching (N=2)                        & 0.50               & 0.323          & 0.567          & 0.2908          & \textbf{0.3249}          & 22.389          \\
DiTFastAttnV2, $\delta=0.4$ (ours) & 0.55               & \textbf{0.238} & \textbf{0.644} &     \textbf{0.2935}      & 0.3236          & \textbf{22.953} \\
DiTFastAttnV2, $\delta=0.6$ (ours) & \textbf{0.63}      & 0.266          & 0.616          & 0.2933 & 0.3246 & 21.842     \\  
\bottomrule
\end{tabular}
\end{table*}

\subsection{Configuration Search Time Comparison}

We compared the compression plan search time between DiTFastAttn and DiTFastAttnV2. For DiTFastAttnV2, we uniformly set the number of calibration prompts to 8. For DiTFastAttn's 1K image generation, we used 6 calibration prompts. For 2K generation, we encountered an Out of Memory (OOM) error when setting the number of calibration prompts to 6, so we present the results with the number of calibration prompts set to 2. In the 2K image generation task, the compression plan search for DiTFastAttn took over 10 hours, whereas our approach required only 15 minutes. This significantly reduces the cost of hyperparameter tweaking.


\begin{figure}[]
    \includegraphics[width=1\linewidth]{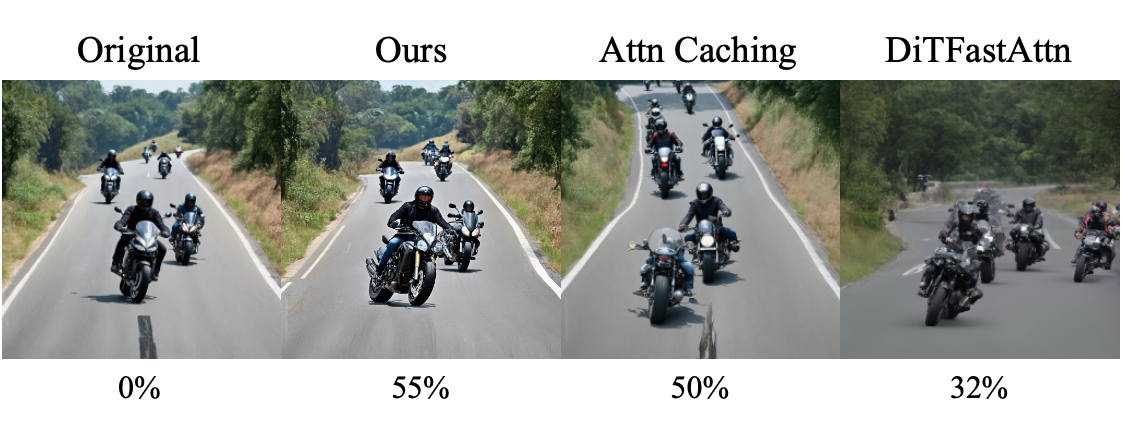}
    \caption{Visual comparison of different attention compression methods. DiTFastAttnV2 maintains the visual quality while incomplete objects and artifacts found in the generation result from other methods.}
    \label{fig:method_example}
\end{figure}

\subsection{FLOPs Reduction and Speedup}

We implement our kernel based on FlashAttention2~\cite{dao2023flashattention}, reducing all unnecessary operations that are not needed in our use cases. We conduct a comprehensive performance evaluation comparing our Head-wise Multi-strategy Attention kernel against the FlexAttention implementation across various configurations on a single NVIDIA A100 GPU. Our experiments employed two distinct token quantity settings: 4096 vision tokens with 512 text tokens (corresponding to 1K image generation), 
16384 vision tokens with 512 text tokens (corresponding to 2K image generation).

For each configuration, we evaluated attention map sparsity levels of 25\%, 50\%, and 75\%. Table~\ref{tab:speed_result_compare} presents the kernel latency and speedup measurements relative to the FlashAttention full attention baseline.


\begin{table}[]
\centering
\caption{Speed comparison among Headwise arrow attention, FlexAttention and FlashAttention2 Our data format is organized as \textit{latency/ms (speedup)} and the speedup is compared with FlashAttention2}
\label{tab:speed_result_compare}
\resizebox{0.85\linewidth}{!}{ 

\begin{tabular}{@{}cccc@{}}
\toprule

\multicolumn{4}{c}{Visual Tokens: 4096 \quad Language Tokens: 512}\\
\multicolumn{4}{c}{FlashAttention2 Full Attention Time: 5.87(ms)}\\  \midrule
              & 25\%                            & 50\%                 & 75\%                 \\ 
Flex                       & 5.08(1.16$\times$)                         & 3.45(1.70$\times$)            & 1.89(3.08$\times$)         \\
ours                       & 4.38(1.34$\times$)                         & 3.08(1.91$\times$)            & 1.81(3.23$\times$)         \\ \midrule
\multicolumn{4}{c}{Visual Tokens: 16384 \quad Language Tokens: 512}\\
\multicolumn{4}{c}{FlashAttention2 Full Attention Time: 9.81(ms)}\\  \midrule
              & 25\%                            & 50\%                 & 75\%                 \\ 
Flex                       & 8.05(1.22$\times$)                  & 5.82(1.69$\times$)                   & 3.18(3.09$\times$)         \\
ours                       & 6.81(1.44$\times$)                  & 4.99(1.96$\times$)                   & 2.76(3.55$\times$)         \\ \midrule
\end{tabular}
}
\end{table}

From the result, we can see that, in all test cases, headwise arrow attention is much faster than FlexAttention and lives up to or even surpasses the ideal speedup ratio.

When attention map sparsity is 25\%, 50\%, 75\%, the ideal speedup is roughly  $1.33\times$, $2\times$ and $4\times$. At lower computational loads (75\% sparsity), both Head-wise Arrow Attention and FlexAttention fall slightly short of the ideal acceleration. For Head-wise Arrow Attention, this is attributable to the computational overhead required for determining the arrow pattern for each thread block in the GPU architecture. With a small sparsity, this overhead noticeably impacts overall performance.
However, as the sparsity and computational workload increase (particularly at 25\% sparsity with larger token counts), Head-wise Arrow Attention not only approaches but occasionally exceeds the theoretical speedup limits. This superior performance stems from two key advantages: The block-granularity computing approach requires only a single masking operation for attention scores. The proportional impact of the overhead diminishes as the total computation increases.

\begin{figure}[]
    \includegraphics[width=1\linewidth]{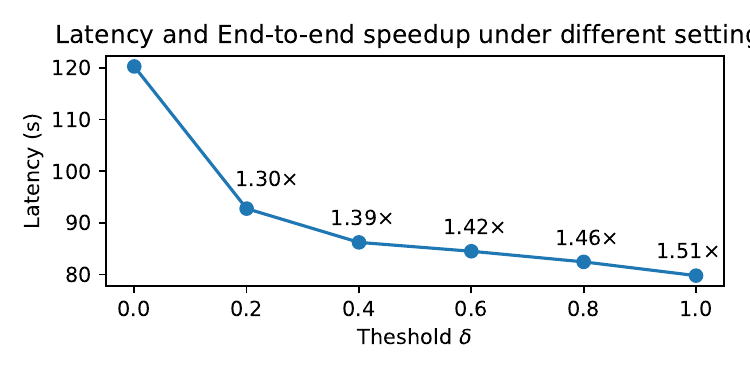}
    \caption{Latency and End-to-end Speedup of DiTFastAttnV2 for FLUX 2K Image Generation}
    \label{fig:latentcy_speedup}
\end{figure}

As shown in Fig~\ref{fig:latentcy_speedup}, with the fused Kernel, for FLUX 2K image generation, DiTFastAttnV2 achieves up to 1.5$\times$ end-to-end speedup.

\subsection{Ablation Study}


\begin{table}[]
\caption{Generation Result of Stable Diffusion 3 1K images with Different Method Set. AA and OC stand for Arrow Attention and Output Caching, respectively.}
\label{tab:ablation_method}
\centering
\resizebox{\linewidth}{!}{ 
\begin{tabular}{lllll}
\toprule
Method Set & \begin{tabular}[c]{@{}l@{}}Attention \\ Sparsity\end{tabular} & LPIPS & SSIM & HPSv2 \\ \hline
                  & 0    & -     & -     & 0.2926 \\
AA & 0.30 & 0.275 & 0.608 & 0.2943 \\
AA \& OC  & 0.55 & 0.238 & 0.644 & 0.2935 \\
w/ CFG Sharing     & 0.54 & 0.249 & 0.649 & 0.2913 \\
w/ Residual Sharing     & 0.56 & 0.196 & 0.704 & 0.2906 \\ \hline
\end{tabular}
}
\end{table}

\subsubsection{Candidate Method Selection}
We experimented with the impact of adding different methods on the generation effectiveness of DiTFastAttnV2. While keeping the $\delta=0.4$ unchanged, we incrementally tried adding various methods including Arrow Attention, and Output Caching, and CFG sharing into the method set. For each combination of methods, we generated 5000 images to evaluate the impact on the model's attention reception and generation effectiveness, as illustrated in the figure. Using only ArrowAttention, the model achieved a 40\% attention sparsity with an HPSv2 score comparable to that of the original model's generated images. When we added output caching to the method set, we achieved a 10\% improvement in attention sparsity at the same threshold, although the HPSv2 score slightly decreased, the generated results were closer to those of the original model. Building on this, we further add CFG sharing and residual sharing that is proposed in DiTFastAttn. We found that applying the CFG sharing method to attention did not significantly improve attention sparsity. We believe that the design of MMDiT eliminates the need for CFG. Therefore, even though CFG can still be applied, the compression potential CFG sharing offers due to redundancy is not actually significant. After adding residual sharing, images generated by the model are structurally more similar to original images but have a drop in HPSv2 score, indicating a slight drop in aesthetic quality. Thus, we leave CFG sharing and arrow attention with residual sharing as options, but when forming the default method set, we select only arrow attention and output caching. 

\subsubsection{Head Constraint Coefficient}


We compared the generation results of attention sparsity under different constraint coefficients on the 1K image generation task of Stable Diffusion 3. We tested the SSIM and LPIPS of images generated with c=1 (fixed threshold for all heads), c=1.5, c=2, and without any constraints, against the original generated images. We found that the model achieved the lowest LPIPS (0.238) and highest SSIM (0.644) when the constraint coefficient was set to 1.5. Further relaxation of the constraint coefficient might lead to a decrease in similarity. Based on this finding, we set c=1.5 as the default constraint coefficient and used this coefficient to generate the experimental results presented in the other parts of this paper.

\begin{table}[]
\centering
\caption{SD3 1K Image Generation Similarity under Different Constraint Coefficient}
\label{tab:cc_study}
\begin{tabular}{lccc}
\toprule
\begin{tabular}[c]{@{}l@{}}Constraint\\ Coefficient\end{tabular} & \begin{tabular}[c]{@{}l@{}}Attention\\ Sparsity\end{tabular} & LPIPS & SSIM \\ 
\midrule
-                      & 0.50          & 0.249 & 0.640 \\
$c=1$                  & 0.55          & 0.240 & 0.641 \\
$c=1.5$                & 0.55          & \textbf{0.238} & \textbf{0.644} \\
$c=2$                  & 0.55          & 0.253 & 0.627 \\
\bottomrule
\end{tabular} 
\end{table}


\section{Conclusion}
In this paper, we propose an efficient approach to accelerating multi-modal diffusion transformers from a head-specific perspective. By leveraging head-wise attention patterns and a novel caching mechanism, our method achieves a high compression ratio in theoretical acceleration metrics such as FLOPs. Furthermore, our efficient kernel design for head-specific attention enables actual speedup on deployment hardware. Additionally, we significantly reduce the search time, achieving optimization within minutes. Extensive experiments demonstrate the effectiveness and practicality of our approach across various benchmarks.

{
    \small
    \bibliographystyle{ieeenat_fullname}
    \bibliography{example_paper}
}

\end{document}